\documentclass[10pt,twocolumn,letterpaper]{article}

\usepackage{iccv}              
\newcommand\red[1]{\textcolor{red}{#1}}

\definecolor{iccvblue}{rgb}{0.21,0.49,0.74}
\usepackage[pagebackref,breaklinks,colorlinks,allcolors=iccvblue]{hyperref}
\usepackage{graphicx}
\usepackage{amsmath}
\usepackage{amssymb}
\usepackage{booktabs}

\usepackage{mathtools}
\usepackage{amsmath}
\usepackage{caption}
\usepackage{subcaption}
\usepackage{wrapfig}
\usepackage{multirow}
\usepackage{amssymb}
\usepackage{pifont}
\usepackage{enumitem}
\usepackage{xcolor}
\usepackage{graphicx}
\usepackage{bbm}
\usepackage{mathrsfs}
\usepackage{stfloats}

\usepackage{amsmath,amsfonts,bm}

\def\eqref#1{equation~\ref{#1}}

\def\1{\bm{1}}

\DeclareMathAlphabet{\mathsfit}{\encodingdefault}{\sfdefault}{m}{sl}
\SetMathAlphabet{\mathsfit}{bold}{\encodingdefault}{\sfdefault}{bx}{n}

\def\paperID{3043} 
\def\confName{ICCV}
\def\confYear{2025}

\title{SummDiff: Generative Modeling of Video Summarization with Diffusion}

\author{
Kwanseok Kim\textsuperscript{1}\thanks{Equal contribution}, Jaehoon Hahm\textsuperscript{2}\footnotemark[1], Sumin Kim\textsuperscript{1}, Jinhwan Sul\textsuperscript{3}, Byunghak Kim\textsuperscript{4}, Joonseok Lee\textsuperscript{1}\thanks{Corresponding author} \\
\textsuperscript{1}Seoul National University, \textsuperscript{2}UIUC, \textsuperscript{3}Georgia Institute of Technology, \textsuperscript{4}Hyundai Card \\
{\tt\small kjvd1009@snu.ac.kr, jh141@illinois.edu, sumink@snu.ac.kr,} \\
{\tt\small jsul7@gatech.edu, byunghak.kim@hcs.com, joonseok@snu.ac.kr}
}

\begin{document}
\maketitle
\begin{abstract}

Video summarization is a task of shortening a video by choosing a subset of frames while preserving its essential moments. Despite the innate subjectivity of the task, previous works have deterministically regressed to an averaged frame score over multiple raters, ignoring the inherent subjectivity of what constitutes a ``good'' summary. We propose a novel problem formulation by framing video summarization as a conditional generation task, allowing a model to learn the distribution of good summaries and to generate multiple plausible summaries that better reflect varying human perspectives.
Adopting diffusion models for the first time in video summarization, our proposed method, SummDiff, dynamically adapts to visual contexts and generates multiple candidate summaries conditioned on the input video. Extensive experiments demonstrate that SummDiff not only achieves the state-of-the-art performance on various benchmarks but also produces summaries that closely align with individual annotator preferences.
Moreover, we provide a deeper insight with novel metrics from an analysis of the knapsack, which is an important last step of generating summaries but has been overlooked in evaluation.
\end{abstract}


\section{Introduction}
\label{sec:intro}



Recently, short-form videos draw significant attention on video sharing platforms, with a trend that consumers increasingly prefer to quickly grasp the content.
They often compressively convey contents that are originally in a longer form, summarizing the core contents into a shorter one; \emph{e.g.}, sport games highlights or movie summarization.
This task of selecting core parts of a long video to construct a shorter one 
is called \textit{video summarization}.
This task is inherently subjective, since there can be multiple criteria for a `good summary'; \emph{e.g.}, comprehensively covering the entire storyline or subjectively selecting impressive parts of the video (highlight detection).
Due to this inherent subjectivity, most video summarization or highlight detection datasets~\cite{gygli2014creating,song2015tvsum} offer annotations by multiple raters to reflect various perspectives.



Since each annotator may have different opinion on the importance of a frame, most existing methods~\cite{apostolidis2021combining, fajtl2019summarizing, jiang2022joint, xu2021cross, son2024csta, he2023align} take the frame-level importance scores averaged across multiple annotators as their target label, and are trained to predict them.
This frame-level score aggregation looks reasonable in some sense, but in fact it loses the various perspectives to summarize each video.
For instance, suppose half of the annotators select clips from the first quarter of the video while the other half select clips from the last quarter. If one simply averages their frame scores, both the first and last quarters end up with similar importance, obscuring the two distinct valid summaries.
That is, this simple regression to the averaged frame-level importance scores fail to preserve multiple viewpoints to summarize the video.

In order to preserve and reflect various views to summarize a video, we pose a \textit{distribution} of its good summaries and let the model to learn it, instead of giving an already-aggregated single ground truth importance score.
Then, the video summarization task can be seen from a generative perspective; \emph{i.e.}, a process of learning and inferring the distribution of good summaries conditioned on the input video.
Specifically, the model is now in charge of estimating the distribution of plausible summaries for the given video.
Once trained, it allows us to sample multiple summaries from the estimated conditional data manifold.
To the best of our knowledge, this generative approach to the video summarization has not been extensively studied, except for a few works~\cite{mahasseni2017unsupervised,apostolidis2020ac} that applied adversarial losses to construct a summary that looks like the original video.

Formulating the video summarization problem as a conditional generation task, we propose to adopt the generative diffusion~\cite{song2020score, ho2020denoising} mechanism, which has been successfully applied to various conditional generation tasks \cite{rombach2022high,ho2022classifier,nichol2021glide}.
Specifically, conditioned on the input video, our proposed \textbf{SummDiff} model learns to denoise a random importance score vector over the given video into an importance score vector sampled from the distribution that corresponds to a good summary of the video.
In contrast to the previous deterministic methods, our approach allows to sample multiple plausible summaries for a given video starting from a different random noise, better aligned with the subjective nature of summarization where we usually have various true labels reflecting multiple views.

Extensive experiments demonstrate that our model outperforms existing baselines across multiple datasets.
Also, we revisit the knapsack, a relatively unexplored step in the summarization evaluation in spite of its nontrivial impact on the performance, and propose additional novel metrics based on this analysis to provide deeper insights.

Our contributions can be summarized as follows:
\begin{itemize}[leftmargin=5mm]
    \setlength{\itemsep}{0pt}
    \setlength{\parskip}{0pt}
    \item We propose a novel generative viewpoint of video summarization, better suited for the subjective nature of the task, allowing multiple plausible summaries for a video.

    \item We innovatively apply diffusion to the video summarization for the first time, integrating learning the distribution of good summaries into the model.
    

    \item We analyze the knapsack optimization process and propose additional metrics to quantify the optimality of the predicted importance scores.
    
\end{itemize}


\begin{figure*}
  \centering
  \begin{minipage}{.64\linewidth}
    \includegraphics[width=1\textwidth]{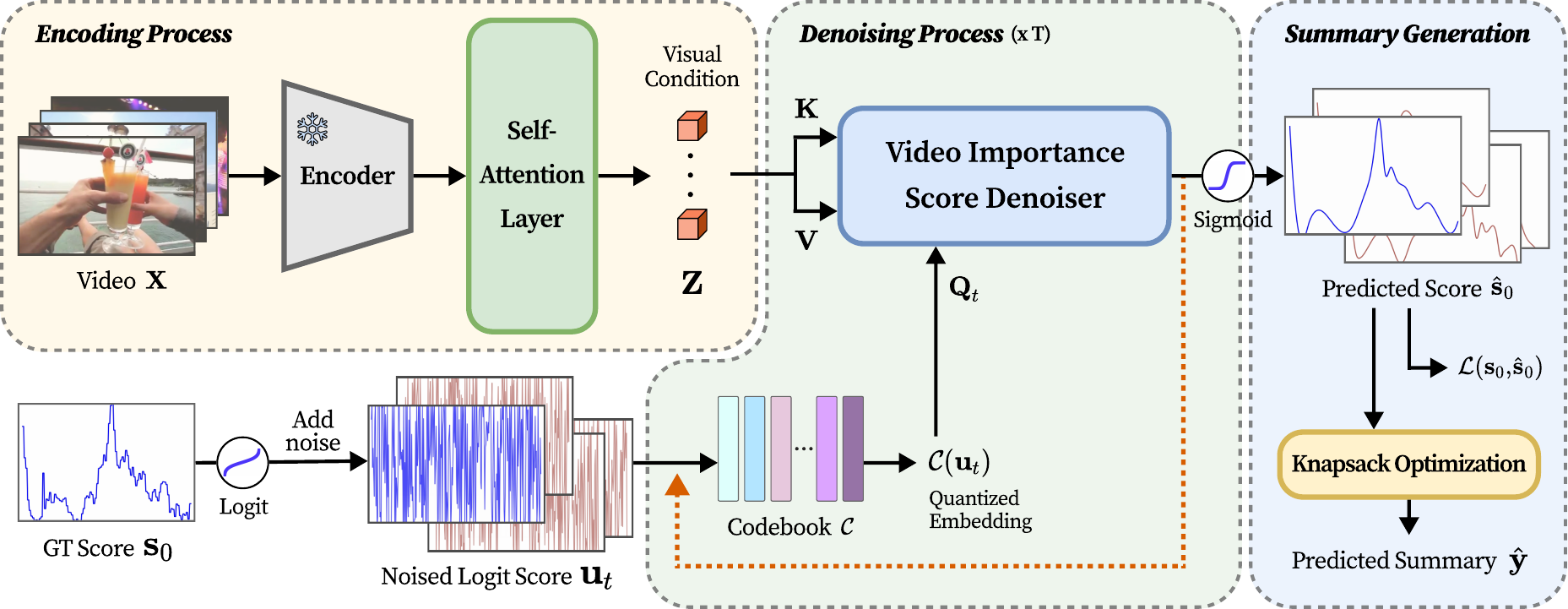}
    \caption{\textbf{Overview of SummDiff.} Given an input video, SummDiff generates importance scores conditioned on video frames. $T$ denotes the number of DDIM steps.}
    \label{fig:overview}
  \end{minipage}
  \hfill
  \begin{minipage}{.33\linewidth}
    \vspace{-0.5cm}\includegraphics[width=\textwidth]{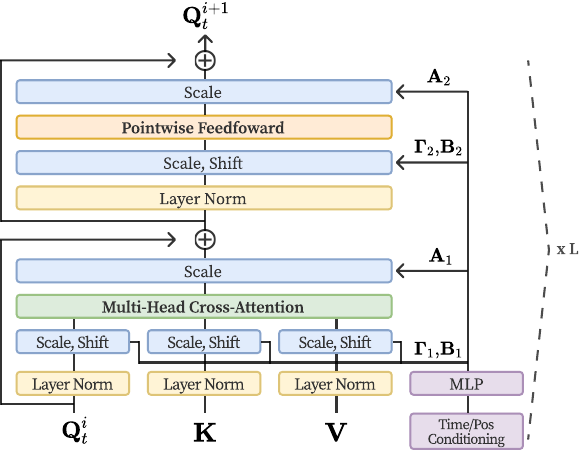}
    \caption{\textbf{Video Importance Score Denoiser}. We use AdaLN layer to inject time and positional conditions, following \cite{peebles2023scalable}.} 
    \label{fig:denoiser}
  \end{minipage}
  \vspace{-0.3cm}
\end{figure*}

\section{Problem Formulation}
\label{sec:prelim:problem}
Given a video of $N$ frames, the objective of video summarization is to identify and select $S < N$ frames that effectively encapsulate the essence of the video content.
Let $\mathbf{X} \equiv \{\mathbf{X}_i \in \mathbb{R}^{H \times W \times 3} : i=0, ..., N-1\} \in \mathbb{R}^{N \times H \times W \times 3} $ be a video, where $H, W$ denote the size of the frames.

A video summary, $\mathbf{y} \equiv \{y_i \in \{0, 1\} : i=0, ..., N-1, \sum_i y_i \le S\} \in \{0,1\}^N$, indicates inclusion (1) or exclusion (0) of each frame.
When a video is provided with multiple ground truth annotations, we denote each individual 
score by $\mathbf{s}^{(r)} \in [0, 1]^N$ and the corresponding binary summary by $\mathbf{y}^{(r)} \in \{0, 1\}^N$ which is obtained following the procedure explained in \cref{sec:method:summary}.
The predicted summary is denoted by $\hat{\mathbf{y}} \equiv \{\hat{y}_i \in \mathbb{R} : i=0, ..., N-1\} \in \mathbb{R}^N$.

Previous models~\cite{apostolidis2021combining,fajtl2019summarizing,son2024csta} have approached video summarization as a regression task, aiming to predict the importance score $\mathbf{s} \in [0, 1]^N$ for each video, often the average of multiple importance scores, $\frac{1}{|R|} \sum_{r \in \mathcal{R}}\mathbf{s}^{(r)}$.
In contrast, our approach allows multiple summaries for each video, aiming to learn the distribution of its plausible summaries.
The previous scheme can be seen as a special case of ours, where every video has a single golden way of summarization, pointed at $\mathbf{s}$ with zero variance.
Under our extended setting, multiple importance scores $\{\mathbf{s}^{(r)} | r \in \mathcal{R} \}$ can be given to the same conditioning video $\mathbf{X}$, forming a probability distribution of plausible importance scores.

\section{Diffusion-based Video Summarization}
\label{sec:method}

Posing the video summarization as a conditional generation task, we introduce our SummDiff model, designed to adapt the distribution of individual importance scores for a given video by learning to denoise.

\subsection{Overall Flow of the Proposed Method}
\label{sec:method:overview}

\cref{fig:overview} illustrates the overall flow of our SummDiff model.

\vspace{0.1cm} \noindent
\textbf{Encoding Process.}
We first encode each frame $\mathbf{X}_i$ for $i = 1, ..., N$ from the input video $\mathbf{X}$ using a pre-trained image encoder.
The extracted features are further contextualized through self-attention~\cite{vaswani2017attention}, as seen in \cref{fig:overview}.
We denote the encoded feature for each individual frame by $\mathbf{z}_i \in \mathbb{R}^D$, and collectively the entire feature matrix by $\mathbf{Z} \in \mathbb{R}^{N \times D}$.

\vspace{0.1cm} \noindent
\textbf{Denoising Process.}
We then learn to denoise an individual importance score vector from a random noise, conditioned on the visual embeddings.
First, the \textit{forward process} adds noise to the ground truth individual importance score
$\mathbf{s}_0 \equiv \mathbf{s}^{(r)} $ and sample a noised one $\mathbf{s}_t = \sqrt{\bar{\alpha}_t} \mathbf{s}_0 + \sqrt{1-\bar{\alpha}_t} \boldsymbol{\epsilon}_t$, where $\boldsymbol{\epsilon}_t \sim \mathcal{N}(0, \mathbf{I}), \bar{\alpha}_t = \prod_{\tau=1}^t(1-\beta_\tau)$, and $t$ is the diffusion time step.
The perturbation kernel at $t$, defined as $q(\mathbf{s}_t | \mathbf{s}_{t-1}) = \mathcal{N}(\mathbf{s}_t ; \sqrt{1-\beta_t} \mathbf{s}_{t-1}, \beta_t \mathbf{I})$, where $\beta_t$ is defined by the variance schedule.
Since $\mathbf{s}_0$ is bounded within $[0, 1]$, it is not straightforward to add Gaussian noise directly to it so we first transform it to its logit, $\mathbf{u}_0 = \log  \frac{\mathbf{s}_0}{1 - \mathbf{s}_0} \in \mathbb{R}^N$, and perform the noising process in this logit space.
For numerical stability, we clip $\mathbf{s}_0$ to $[\epsilon, 1-\epsilon]$, where $\epsilon$ is a small constant.

Then, the \textit{reverse process} progressively removes noise from $\mathbf{s}_t$ to $\mathbf{s}_0$, formulated by $p_\theta(\mathbf{s}_{t-1} | \mathbf{s}_t) = \mathcal{N}(\mathbf{s}_{t-1}; \boldsymbol{\mu}_\theta(\mathbf{s}_t, t), \sigma_t^2 \mathbf{I})$, where $\sigma_t^2$ is determined in relation to $\beta_t$, and the posterior mean $\boldsymbol{\mu}_\theta(\mathbf{s}_t, t)$ is modeled with a trainable neural network.
In this paper, we extend this estimator of posterior mean to be conditioned on a video, similarly to the conditional generation on diffusion models.

Once trained, our denoiser is able to recover a plausible importance score for a video from a random noise, which is used to generate a summary (\cref{sec:method:summary}).
Repeated generation across the noise distribution would converge to the true distribution of plausible scores for the video.




\subsection{Video Importance Score Denoiser}
\label{sec:method:visd}

We formulate the key component of our method, the video importance score denoiser in \cref{fig:denoiser}.
Starting from a noised logit score vector $\mathbf{u}_t \in \mathbb{R}^N$, it predicts a plausible importance score, conditioned on the given video.
Denoted by $f_\theta(\mathbf{u}_t, t, \textbf{Z})$, it learns to denoise the given score vector $\mathbf{u}_t$ at the diffusion time step $t$ under the visual condition $\mathbf{Z}$, producing the denoised score $\hat{\mathbf{u}}_{t-1}$ by one time step, where $\theta$ is the set of its learnable parameters.

\vspace{0.1cm}
\noindent\textbf{Transformer-based Diffusion.}
We denoise the logit-transformed score $\mathbf{u}_t$ given $t$ and $\mathbf{Z}$, predicting $\hat{\mathbf{u}}_{t-1}$, using a transformer-based cross-attention \cite{vaswani2017attention}. 
The logit-transformed score $\mathbf{u}_t$ acts as the query, and the visual condition $\mathbf{Z}$ is used as key/value.
This setup allows the model to effectively denoise $\mathbf{u}_t$ conditioned on the information from the input video, ensuring consistency between the predicted logit score $\hat{\mathbf{u}}_{t-1}$ and the condition $\mathbf{Z}$.

To apply dot-product cross-attention, the dimensionality should match for queries and keys.
To this end, we \textit{quantize} the importance scores into a predefined number ($K$) of uniform segments.
Specifically, we convert the noised logit $\mathbf{u}_t$ back to its original bounded range $[0, 1]$ by applying sigmoid $1/(1+e^{-\mathbf{u}_t})$, and divide them into $K$ equally-binned segments.
Each score range is associated with a learnable embedding of size $D$ (codebook in \cref{fig:overview}).
Based on the codebook $\mathcal{C}$, we map the scores $\mathbf{u}_t \in \mathbb{R}^N$ to their corresponding quantized embedding $\mathcal{C}(\mathbf{u}_t) \in \mathbb{R}^{N \times D}$.
Our denoiser $f_\theta(\mathcal{C}(\mathbf{u}_t), t, \mathbf{Z})$ takes this $\mathcal{C}(\mathbf{u}_t)$ as input, instead of $\mathbf{u}_t$.
See \cref{sec:exp:ablation} for anaylsis on the size of codebook.

\vspace{0.1cm}
\noindent\textbf{Query Formation in Cross-Attention.}
Following \cite{ho2020denoising}, the time step $t$ is embedded as $\boldsymbol{\tau} \in \mathbb{R}^D$ using sinusoidal functions and an MLP.
Considering the sequential nature of videos, we also introduce temporal positional embeddings $\boldsymbol{\Phi}\in \mathbb{R}^{N \times D}$ using sinusoidal functions. 

The simplest way to set the query in the cross-attention would be to add all inputs except for the condition $\mathbf{Z}$; that is, $\mathbf{Q}_t = \mathcal{C}(\mathbf{u}_t) + 
\boldsymbol{\tau} + \boldsymbol{\Phi}$.
However, we leverage AdaLN-Zero block~\cite{peebles2023scalable} to integrate the time embedding $\boldsymbol{\tau}$ and temporal positional encoding $\boldsymbol{\Phi}$ more effectively by separating them from the query $\mathbf{Q}_t$, preventing information mixing. Hence, the query becomes $\mathbf{Q}_t = \mathcal{C}(\mathbf{u}_t)$ and $\boldsymbol{\tau}, \boldsymbol{\Phi}$ are conditioned via scale-shift operation.
See \cref{sec:exp:ablation} for ablation studies.



We conduct cross-attention with $\mathbf{Q}_t$ as queries and visual condtions $\mathbf{Z} \in \mathbb{R}^{N \times D}$ as keys $\mathbf{K}$ and values $\mathbf{V}$.
We take an MLP from the AdaLN output to regress the scale and shift parameters, denoted by $\mathbf{A}_1$, $\mathbf{B}_1$, $\mathbf{\Gamma}_1$, $\mathbf{A}_2$, $\mathbf{B}_2$, and $\mathbf{\Gamma}_2 \in \mathbb{R}^{N \times D}$.
As depicted in \cref{fig:denoiser}, they scale and shift $\mathbf{Q}_t$, $\mathbf{K}$, and $\mathbf{V}$.
Then, they are passed through cross-attention with skip connections and a subsequent rescaling.
Formally,
\begin{align}
    {\mathbf{H}^i}^{_{'}} &= \mathbf{\Gamma}_1 \odot {\mathbf{H}^i} + {\mathbf{H}^i} + \mathbf{B}_1 \nonumber \\
    \mathbf{X}_1 &= \mathbf{A}_1 \odot \text{softmax}({\mathbf{Q}_t^i}' {\mathbf{K}^i}'^\top) {\mathbf{V}^i}' + {\mathbf{Q}_t^i}' \nonumber \\
    \mathbf{X}_2 &= \mathbf{\Gamma}_2 \odot \mathbf{X}_1  + \mathbf{X}_1 + \mathbf{B}_2 \nonumber \\
    \mathbf{Q}_t^{i+1} &= \mathbf{A}_2 \odot \text{MLP}(\mathbf{X}_2) + \mathbf{X}_2, \nonumber 
    \label{eq:denoiser}
\end{align}
where ${\mathbf{H}^i}^{(_{'})} \in \{ {\mathbf{Q}_t^i}^{(_{'})}, {\mathbf{K}^i}^{(_{'})}, {\mathbf{V}^i}^{(_{'})} \}$ denotes the matrices used for attention,
$\odot$ is the (broadcasted) Hadamard product, and $i = 1, ..., L$ is the layer index.

\vspace{0.1cm}
\noindent\textbf{Training.}
We train the denoiser $f_\theta$ to estimate the true importance score $\hat{\mathbf{s}}_{0}$ after acting upon a fully-connected layer and a sigmoid function $\sigma$.
We minimize the following loss \cite{chen2022analog} on each annotator's individual importance score:
\begin{equation}
  \mathcal{L}(\mathbf{s}_0, \hat{\mathbf{s}}_0) = \|\mathbf{s}_0-\hat{\mathbf{s}}_0\|_2^2
  = \| \mathbf{s}_0 - \sigma(\text{FC}(f_{\theta}(\mathcal{C}(\mathbf{u}_t), t, \mathbf{Z}))) \|_2^2. \nonumber
  \label{loss}
\end{equation}
The learnable embeddings in the codebook $\mathcal{C}$ are compositely optimized, finding an effective representation for $\mathbf{u}_t$ during training.


\vspace{0.1cm}
\noindent\textbf{Inference.}
Our model generates a logit-transformed importance score from a random noise $\mathbf{u}_T \sim \mathcal{N}(0, \mathbf{I})$ for a given video.
Employing the reverse diffusion process \cite{song2020denoising},
it iteratively refines the logit score towards cleaner estimations:
\begin{align}
  \hat{\mathbf{u}}_{t-1} &= \sqrt{1-\bar{\alpha}_{t-1}-\sigma_t^2}\frac{\hat{\mathbf{u}}_t-\sqrt{\bar{\alpha}_t}f_{\theta}(\mathcal{C}(\mathbf{u}_t), t, \mathbf{Z}))}{\sqrt{1-\bar{\alpha}_t}} \nonumber \\
  &+ \sqrt{\bar{\alpha}_{t-1}}f_{\theta}(\mathcal{C}(\mathbf{u}_t), t, \mathbf{Z})) + \sigma_t\epsilon_t \in \mathbb{R}^N, \nonumber
  \label{eq:inference}
\end{align}
where $\sigma_t = \sqrt{(1 - \bar{\alpha}_{t-1}) / (1 - \bar{\alpha}_{t})} \sqrt{1- \bar{\alpha}_t / \bar{\alpha}_{t-1}}$ (DDPM reverse process).
When $t=1$ at the last step, $\hat{\mathbf{u}}_0 = f_{\theta}(\mathcal{C}(\mathbf{u}_1), 1, \mathbf{Z})$ is directly used to remove stochasticity at the end of inference process. Finally, we take sigmoid to $\hat{\mathbf{u}}_0$ to derive the final importance score $\hat{\mathbf{s}}_0 \in (0, 1)^N$.

\subsection{Summary Generation} 
\label{sec:method:summary}

From the raw importance score $\hat{\mathbf{s}}_0$,
we apply the standard knapsack-based approach to decide which frames to be included in the final summary.
To make this summary video more realistic, it is common to choose at a semantic clip level instead of individual frame level.
We adopt a widely-used Kernel Temporal Segmentation (KTS)~\cite{zhang2016video, potapov2014category} to partition a video into disjoint temporal intervals.
We take the average score among the corresponding frames as the clip-level importance; \emph{i.e.}, $v_i = \sum_{j=t_i}^{t_{i+1}} \hat{\mathbf{s}}_{0,j} / (t_{i+1} - t_i)$ for the $i$-th clip composed of frames from $t_i$ to $t_{i+1}$.
Subsequently, we select the clips based on $v_i$ by solving the binary knapsack problem (KP)~\cite{hifi2005sensitivity} with dynamic programming~\cite{song2015tvsum}:
\begin{equation}
    \mathcal{KP}(\boldsymbol{v}, \boldsymbol{w}, \rho) \equiv 
    \underset{\boldsymbol{s} \in \{0, 1\}^M}{\text{argmax}} 
    \sum_{i=1}^{M} v_i s_i 
    \quad \text{s.t.} \ \sum_{i=1}^{M} w_i s_i \leq \rho N, \nonumber
\end{equation}
where $\boldsymbol{v}, \boldsymbol{w}, \rho$ denotes the values of each clip ($v_i \in [0,1]$), costs of selecting each clip, and the budget constraint ratio (\textit{e.g.}, $\rho=0.15$; 15$\%$ of the video length), respectively. If the argmax has multiple solutions, we will abuse the notation and denote $\mathcal{KP}(\boldsymbol{v}, \boldsymbol{w}, \rho)$ as an arbitrary element of the solution set.
\section{Experiments}
\label{sec:exp}

\subsection{Experimental Settings}
\label{subsec:setting}

\noindent
\textbf{Datasets.}
We evaluate our method on three benchmarks.
TVSum~\cite{song2015tvsum} and SumMe~\cite{gygli2014creating} are traditional datasets, containing 50 and 25 videos respectively and manually labeled annotations by up to 20 raters.
Mr.~HiSum \cite{sul2024mr} is a large-scale 
dataset, composed of 31,892 videos and importance scores derived from YouTube Most Replayed statistics.
It provides an importance score $\mathbf{s}_0$ and a corresponding summary $\mathbf{y}_0$ for each video, averaged over 50,000+ viewers.
We use the default features for each dataset, GoogLeNet~\cite{szegedy2015going} for TVSum and SumMe, and Inception-v3 \cite{szegedy2016rethinking} PCA-ed to 1024D~\cite{abu2016youtube} for Mr.~HiSum.

We adopt two dataset splits. First, we randomly split TVSum and SumMe into 7:1:2 for train, validation, and test set, and report averaged test scores for 5 random splits (namely, TVT).
Also, following prior works~\cite{terbouche2023multi,son2024csta}, we include five-fold cross-validation results (5FCV),
although these may suffer from overfitting to the test set due to the lack of explicit validation set, and thus may not generalize well on unseen videos.
For Mr.~HiSum, we use its original train/validation/test split.
We run all experiments 5 times.

\vspace{0.1cm}
\noindent\textbf{Evaluation Metrics.}
Following recent practice \cite{otani2019rethinking, son2024csta, he2023align, li2023progressive}, we report rank order statistics, Kendall's $\tau$ \cite{kendall1945treatment} and Spearman's $\rho$ \cite{zwillinger1999crc}, 
to asses how well the model ranks frame importance.
F1 score had been widely used in video summarization, but recent literature~\cite{terbouche2023multi,son2024csta} reports that it is excessively sensitive to video segmentation and unreasonably favors summaries composed of many short shots~\cite{otani2019rethinking}, while disregarding longer key shots~\cite{son2024csta}.
As an alternative to F1, we propose additional metrics that represent the influence of importance score through an analysis of the knapsack algorithm in \cref{sec:analysis}.

We further evaluate our method on video highlight detection following~\cite{sul2024mr}.
First, we uniformly divide the input video into 5-second-long shots and calculate the average frame scores for each shot.
The top $\rho \in \{0.15, 0.5\}$ of these shots are designated as ground truth highlights.
We report Mean Average Precision (MAP),
following \cite{hong2020mini, zhang2016video, panda2017collaborative}.
See \cref{sec:impl_detail_appendix} for more implementation details.

\begin{table}[t]
    \centering
    \footnotesize
    \setlength{\tabcolsep}{8pt}
    \begin{tabular}{@{}lcccc@{}}
    \toprule
    \multirow{2}{*}{Method}  & \multicolumn{2}{c}{SumMe} & \multicolumn{2}{c}{TVSum} \\
    \cmidrule(lr){2-3} \cmidrule(lr){4-5}
    & $\tau$ & $\rho$ & $\tau$ & $\rho$ \\
    \midrule
    Random                  & 0.000 & 0.000 & 0.000 & 0.000 \\
    Human                   & 0.205 & 0.213 & 0.177 & 0.204 \\
    \midrule
    A2Summ~\cite{he2023align}           & 0.088 & 0.096 & 0.157 & 0.206 \\
    VASNet~\cite{fajtl2019summarizing}  & 0.089 & 0.099 & 0.153 & 0.205 \\
    PGL-SUM~\cite{apostolidis2021combining} & 0.104 & 0.116 & 0.141 & 0.186 \\
    CSTA~\cite{son2024csta}             & \underline{0.108} & \underline{0.120} & \underline{0.168} & \underline{0.221} \\
    \midrule
    \textbf{SummDiff (Ours)}           & \textbf{0.133} & \textbf{0.148} & \textbf{0.173} & \textbf{0.226} \\
    \bottomrule
    \end{tabular}
    \vspace{-0.2cm}
    \caption{\textbf{Comparison of models trained under TVT (train/val/test split)} on SumMe~\cite{gygli2014creating} and TVSum~\cite{song2015tvsum}. Best and second-best results are \textbf{boldfaced} and \underline{underlined}, respectively.}
    \label{tab:tvt_results}
    \vspace{-0.3cm}
\end{table}

\begin{table}[t]
    \centering
    \footnotesize
    \setlength{\tabcolsep}{8pt}
    \begin{tabular}{@{}lcccc@{}}
    \toprule
    \multirow{2}{*}{Method} & \multicolumn{2}{c}{SumMe} & \multicolumn{2}{c}{TVSum} \\
    \cmidrule(lr){2-3} \cmidrule(lr){4-5}
    & $\tau$ & $\rho$ & $\tau$ & $\rho$ \\
    \midrule
    Random                  & 0.000 & 0.000 & 0.000 & 0.000 \\
    Human                   & 0.205 & 0.213 & 0.177 & 0.204 \\
    \midrule
    DSNet-AF~\cite{zhu2020dsnet}        & 0.037  & 0.046    & 0.113  & 0.138    \\
    DSNet-AB~\cite{zhu2020dsnet}        & 0.051  & 0.059    & 0.108  & 0.129    \\
    SUM-GAN~\cite{mahasseni2017unsupervised}    & 0.049 & 0.066 & 0.024 & 0.031 \\
    AC-SUM-GAN~\cite{apostolidis2020ac}         & 0.102 & 0.088 & 0.031 & 0.041 \\
    CLIP-It~\cite{narasimhan2021clip}           & -     & -     & 0.108 & 0.147 \\
    iPTNet~\cite{jiang2022joint}                & 0.101 & 0.119 & 0.134 & 0.163 \\
    A2Summ~\cite{he2023align}                   & 0.108 & 0.129 & 0.137 & 0.165 \\
    VASNet~\cite{fajtl2019summarizing}          & 0.160 & 0.170 & 0.160 & 0.170 \\
    PGL-SUM~\cite{apostolidis2021combining}     & -     & -     & 0.157 & 0.206 \\
    AAAM~\cite{terbouche2023multi}              & -     & -     & 0.169 & 0.223 \\
    MAAM~\cite{terbouche2023multi}              & -     & -     & 0.179 & 0.236 \\
    VSS-Net~\cite{zhang2023vss}                 & -     & -     & 0.190 & 0.249 \\
    DMASum~\cite{wang2020query}                 & 0.063 & 0.089 & \textbf{0.203} & \textbf{0.267} \\
    SSPVS~\cite{li2023progressive}              & 0.192 & 0.257 & 0.181 & 0.238 \\
    CSTA~\cite{son2024csta}                     & \underline{0.246} & \underline{0.274} & 0.194 & \underline{0.255} \\
    \midrule
    \textbf{SummDiff (Ours)}                    & \textbf{0.256} & \textbf{0.285} & \underline{0.195} & \underline{0.255} \\
    \bottomrule
    \end{tabular}
    \vspace{-0.2cm}
    \caption{\textbf{Comparison of models trained under 5FCV (5-Fold Cross Validation)} on SumMe~\cite{gygli2014creating} and TVSum~\cite{song2015tvsum}. Training under the 5FCV setting tends to overfit the test set.}
    \label{tab:5fcv_results}
    \vspace{-0.3cm}
\end{table}

\begin{table}
    \centering
    \footnotesize
    \renewcommand{\tabcolsep}{4pt}
    {\begin{tabular}{l|ccccc}
    \toprule
    Model & $\tau$ $\uparrow$ &  $\rho$ $\uparrow$ & MAP{\tiny$\rho$ = 50\%} $\uparrow$ & MAP{\tiny $\rho$ = 15\%} $\uparrow$\\
    \midrule
    SUM-GAN \cite{mahasseni2017unsupervised}  & 
    0.067 & 0.095 & 56.62 & 23.56 \\

    VASNet~\cite{fajtl2019summarizing}  &
    0.069 & 0.102 & 58.69 & 25.28 \\

    AC-SUM-GAN \cite{apostolidis2020ac} & 
    0.012 & 0.018 & 55.35 & 21.88 \\

    SL-module \cite{xu2021cross} & 
    0.060 & 0.088 & 58.63 & 24.95 \\

    PGL-SUM~\cite{apostolidis2021combining}  & 
    0.097 & 0.141 & 61.60 & 27.45 \\

    iPTNet ~\cite{jiang2022joint} &
    0.020 & 0.029 & 55.53 & 22.74 \\

    A2Summ ~\cite{he2023align} &
    0.121 & 0.172 & 63.20 & 32.34 \\

    CSTA ~\cite{son2024csta}  &
    \underline{0.128} & \underline{0.185} & \underline{63.38} & \underline{30.42} \\

    \midrule
    
    \textbf{SummDiff (Ours)}  &
    \textbf{0.175} & \textbf{0.238} & \textbf{65.44} & \textbf{33.83} \\
    \bottomrule
    \end{tabular}}
  \vspace{-0.2cm}
  \caption{\textbf{Evaluation on Mr.~HiSum~\cite{sul2024mr}.} See \cref{appendix:full_table} for the full table including standard deviations.}
  \label{tab:new_leaderboard}
  \vspace{-0.5cm}
\end{table}

\subsection{Results and Analysis}
\label{sec:results}


Considering the innate subjectivity of video summarization,
we train our model to learn from each individual score, allowing it to capture multiple ways of summarization for each video.
Specifically, we use 20 individual importance scores per video in TVSum~\cite{song2015tvsum} and 15–18 individual binary summaries in SumMe~\cite{gygli2014creating}, since SumMe does not provide individual importance scores.
\cref{tab:tvt_results} and \ref{tab:5fcv_results} show that SummDiff achieves the best performance on SumMe and TVSum, with the sole exception of TVSum in 5FCV.
While DMA-SUM~\cite{wang2020query} performs the best on TVSum in 5FCV, it significantly underperforms on SumMe, indicating that its high performance on TVSum does not generalize well.
Comparing TVT and 5FCV, most models exhibit a considerable performance gap. 
This demonstrates how much existing models have overfitted to the test set
under the conventional 5FCV protocol.
We adopt TVT as a remedy to this, but due to the extremely small size of these datasets, the reliability of evaluation is still limited \cite{sul2024mr}.

On the larger Mr.~HiSum, we train our model using the aggregated annotation, as it does not provide individual annotations.
This evaluation mainly aims to verify scalability of SummDiff on a large-scale dataset.
\cref{tab:new_leaderboard} presents the performance on Mr.~HiSum, where SummDiff consistently outperforms all baselines across all metrics, significantly surpassing the strongest competitor, CSTA~\cite{son2024csta} with a large margin.
These results highlight scalability and effectiveness of our method even under the single-label setting.

\subsection{Qualitative Demonstration}
\label{sec:method:qual}

\begin{figure}
  \begin{center}
  \vspace{-0.2cm}
  \scalebox{1}[1]{\includegraphics[width=0.42\textwidth]{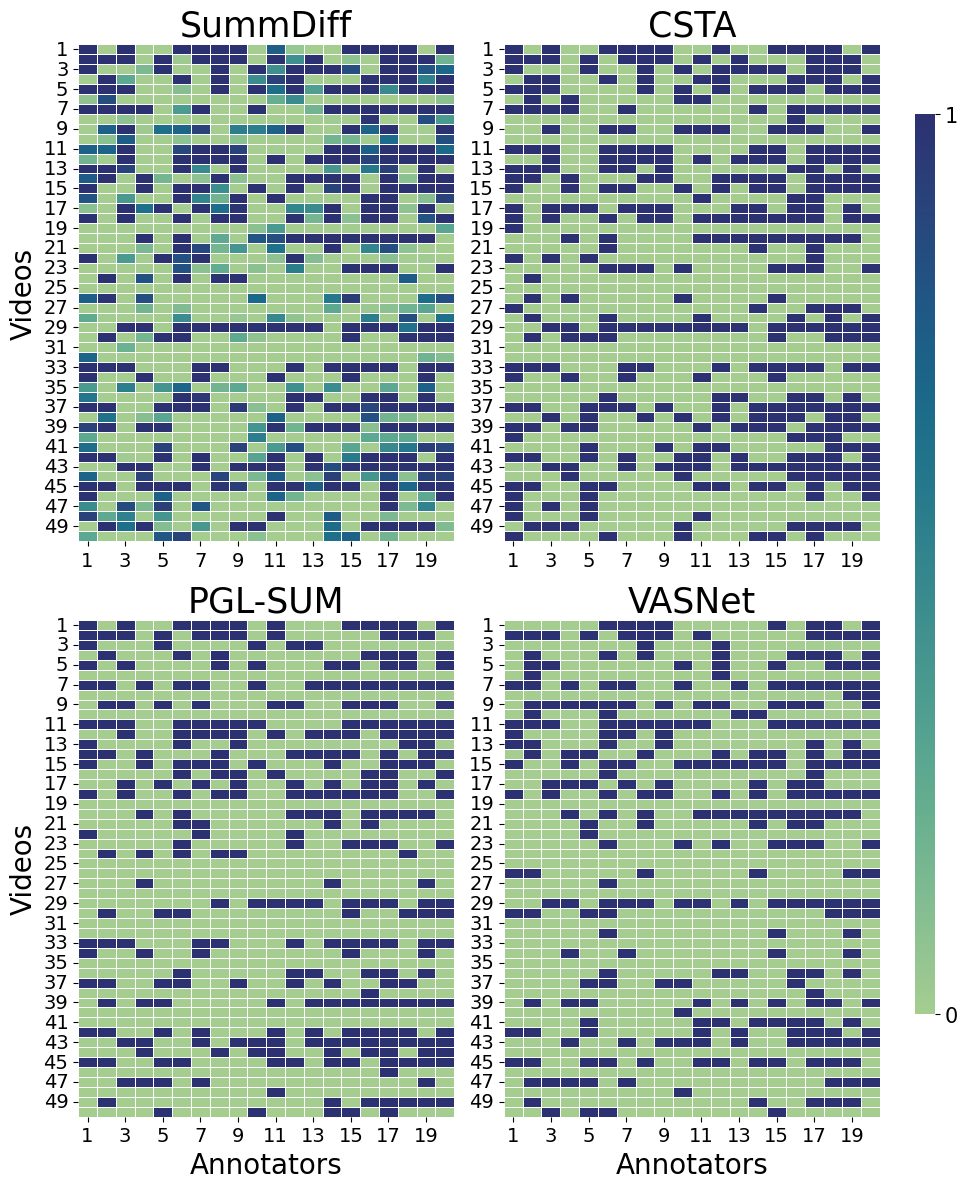}}
  \end{center}
  \vspace{-0.6cm}
  \caption{\textbf{Ratio of summaries with $\tau \ge 0.25$ for each video-annotator pair.} The heatmap illustrates how closely each method’s summary matches with individual annotations. SummDiff covers a larger area of the heatmap, which indicates better coverage over various summaries.}
  \label{fig:heatmap}
  \vspace{-0.5cm}
\end{figure}

\begin{figure}
  \begin{center}
  \includegraphics[width=0.42 \textwidth, clip]{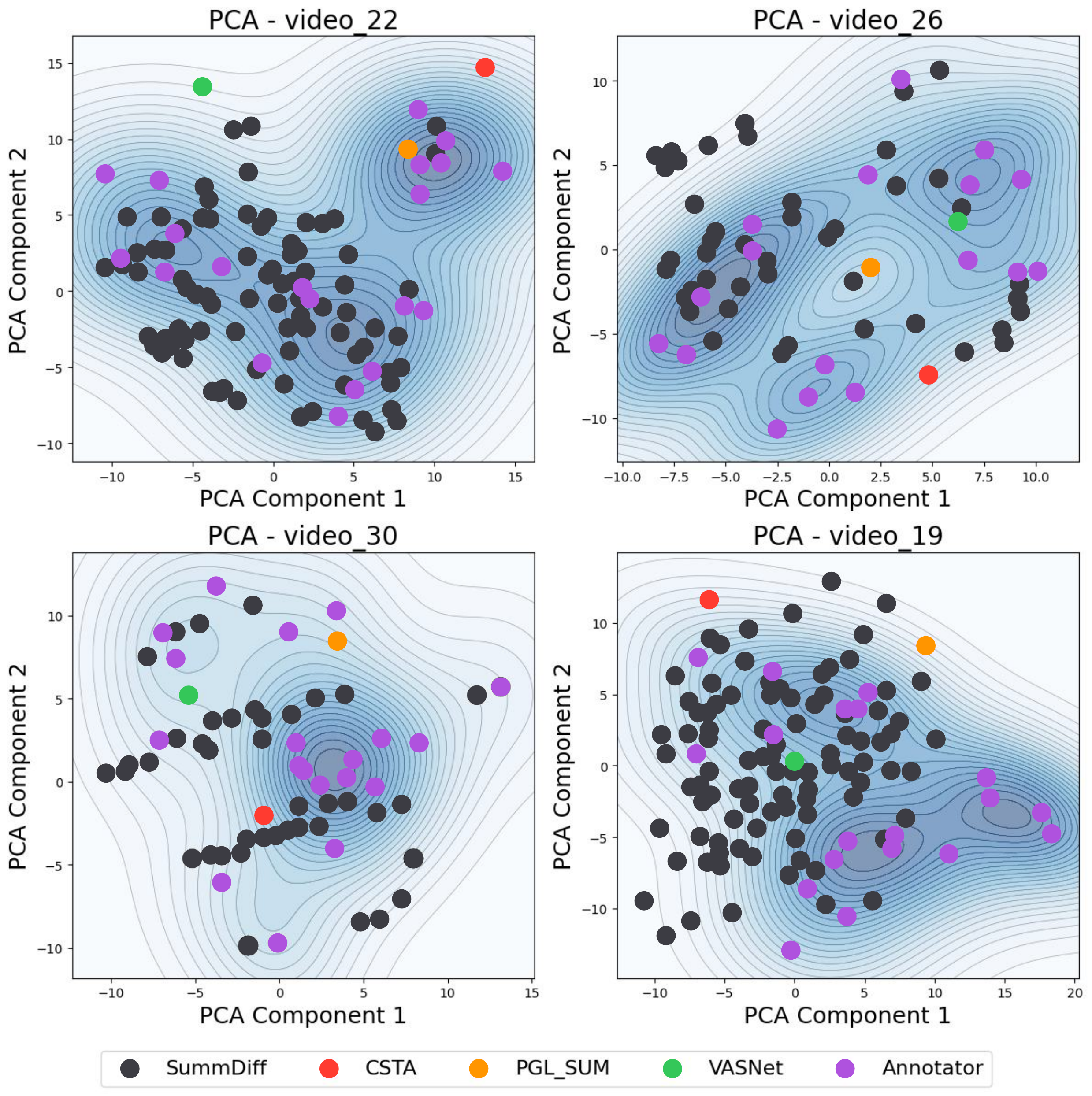}
  \end{center}
  \vspace{-0.6cm}
  \caption{\textbf{Distribution of true and predicted summaries of given videos.} SummDiff can generate various summaries and cover the targeted distribution of summaries, while baselines deterministically predicts a single summary.}
  \label{fig:qualitative}
  \vspace{-0.5cm}
\end{figure}

\begin{figure*}
  \centering
  \includegraphics[width=0.92\textwidth]{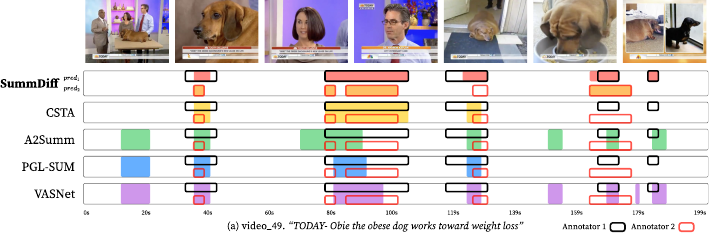}
  \vspace{-0.2cm}
  \caption{\textbf{Demonstration of video summaries generated by competing methods on a TVSum video}. Shaded parts indicate the segments selected by each method, and the two rows of edged boxes within each method indicate the two distinct true annotations.
  The results clearly demonstrate the effectiveness of SummDiff in capturing multiple plausible summaries for a video.
  See another example in \cref{appendix:qual}.}
  \label{fig:demo}
  \vspace{-0.4cm}
\end{figure*}

We demonstrate how our method discovers a variety of possible summaries for a video.
Specifically, we generate video summaries for the 50 videos in TVSum using CSTA, PGL-SUM, VASNet, and ours.
For each generated summary, we measure the Kendall's $\tau$ with all 20 ground truth annotations, respectively.
Considering $\tau\ge 0.25$ as the threshold for the summary to be matched with the annotator, \cref{fig:heatmap} shows if the generated summary for a video (row) matches with each annotator (column).
As the baselines (CSTA, PGL-SUM, and VASNet) summarize a video deterministically, each cell is either completely matched (1) or not (0).
If the summary matches with majority of annotators (\emph{e.g.}, video 29 for CSTA), it means the deterministic summary matches with the dominant way of summarization for the video.
If it matches only with a few annotators (\emph{e.g.}, video 24 for CSTA), it means the generated summary matches with a minor opinion.
For our method, we generate 100 summaries per video from different Gaussian noise vectors, and mark each cell with the ratio of summaries that match with each annotator.
The darker the cell colors are for each row, the more various ways of summaries have been discovered by our method.
Comparing the heat maps, we clearly observe that SummDiff generates significantly more various summaries covering multiple viewpoints, better reflecting the distribution of annotated summaries.


\newcommand{\mycirc}[1][black]{\textcolor{#1}{\ensuremath\bullet}}

We further illustrate the variation and quality of generated summaries in \cref{fig:qualitative}.
The summaries annotated by human raters (\mycirc[Orchid] Annotator) and those generated by ours (\mycirc[darkgray] SummDiff) and baselines (\mycirc[red] CSTA, \mycirc[BurntOrange] PGL-SUM, and \mycirc[Green] VASNet) are projected onto the first two principal components, which are computed using only the human-annotated summaries.
The distribution of true summaries are displayed using a contour map, obtained by Gaussian kernel density estimation.
It demonstrates that our SummDiff generates multiple summaries closely aligned with the distribution of true annotations.
For example, video 22 (top-left) roughly has three ways to summarize, according to the human raters.
SummDiff generates summaries dominantly covering the two modes in the lower-left region, and one case in the upper-right mode.
In contrast, baselines produce a single, less accurate summary, failing to account for the variety inherent in human-annotated summaries.
We observe similar patterns in the other three plots as well.
\cref{fig:qualitative_appendix} provides more examples in \cref{sec:appendix:contour}.

\cref{fig:demo} illustrates example summaries generated by CSTA, A2Summ, PGL-SUM, VASNet, and our SummDiff, for a video selected from TVSum test set.
Shaded parts indicate the segments selected by each method 
(note that SummDiff can generate two different summaries while baselines always produce a single summary),
and the two rows of edged boxes within each method indicate two different true annotations.
These results demonstrate that SummDiff produces more accurate and various summaries, effectively capturing multiple plausible summarizations for each video.

\subsection{Metrics Inspired from Knapsack Optimality}
\label{sec:analysis}


In spite of nontrivial impact of the knapsack at the end of the summary generation, previous evaluation metrics have focused only on the accuracy of the importance scores.
Through a thorough analysis on the knapsack problem (KP), we provide additional metrics that measure the contribution of importance scores.
By accounting for knapsack optimality and clip-level weights, our new metrics are capable of assessing the predicted importance score more accurately than existing metrics such as F1 or ranking-based ones.




\vspace{0.1cm} \noindent
\textbf{Confidence of Importance Score.}
First, we analyze
the conditions under which the same optimal KP solution remains valid despite some perturbations to the profits $\boldsymbol{v}$.
These perturbations can be a modeling to the imperfect estimation of the importance scores in video summarization.

Let $\boldsymbol{y}^*$ be the optimal solution to the original KP (\textit{i.e.}, $\boldsymbol{y}^* = \mathcal{KP}(\boldsymbol{v}, \boldsymbol{w}, \rho)$; see definition in \cref{sec:method:summary}) and $\Gamma \subseteq [N] \equiv \{n \in \mathbb{N} \mid 1 \leq n \leq N\}$ be an arbitrary subset of items associated with the perturbed profits $\boldsymbol{v}'_i = \boldsymbol{v}_i + \Delta v_i$, where $\Delta v_i = 0$ if $i \notin \Gamma$ and $N$ is the number of items.
In our case, $\boldsymbol{v} \in \mathbb{R}^N$ is the ground truth importance scores,
$\boldsymbol{y}^* \in [0, 1]^N$ is the ground truth summary, and the perturbed profit $ \boldsymbol{v}' \in \mathbb{R}^{N}$ corresponds to the imperfectly predicted importance scores.
We further define two disjoint subsets of $\Gamma$.
First, $\Gamma_0^{+} \equiv \{i \in \Gamma \mid y^*_i = 0, \Delta p_i > 0 \}$ refers to the set of items excluded in the original optimal solution but undergoing increased profit, and $\Gamma_1^{-} \equiv \{i \in \Gamma \mid y^*_i = 1, \Delta p_i < 0 \}$ is the opposite, the set of items included in the optimal solution but experiencing a decreased profit.
Then, Hifi \textit{et al.}~\cite{hifi2005sensitivity} claims that the solution $\boldsymbol{y}^*$ remains optimal for the perturbed KP, \textit{i.e.}, $\boldsymbol{y}^* = \mathcal{KP}(\boldsymbol{v}, \boldsymbol{w}, \rho) = \mathcal{KP}(\boldsymbol{v}', \boldsymbol{w}, \rho)$, if
\begin{equation}
    \sum_{i\in \Gamma_0^+} \Delta v_i - \sum_{i\in \Gamma_1^-} \Delta v_i
    \le
    \boldsymbol{v} \cdot \mathcal{KP}(\boldsymbol{v}, \boldsymbol{w}, \rho) - \max(\Psi^+, \Psi^-), \nonumber
\end{equation}
where $\Psi^+ \equiv \max_{i \in \Gamma_0^+}[ \boldsymbol{v} \cdot \mathcal{KP}(\boldsymbol{v} - v_i \hat{e}_i, \boldsymbol{w}, \rho - w_i/N)]$ is the maximum value of the KP when removing any item from $\Gamma_0^+$ and adjusting the constraint, and similarly, $\Psi^- \equiv \max_{i \in \Gamma_1^-}[ \boldsymbol{v} \cdot \mathcal{KP}(\boldsymbol{v} - v_i \hat{e}_i, \boldsymbol{w}, \rho)]$ is the maximum value of the KP when removing any item from $\Gamma_1^-$.

Based on this theorem we propose a new metric, \textbf{Confidence of Importance Score (CIS)}, to estimate how close the predicted scores are to the ground truth:
\begin{equation}
    \text{CIS} = \sum_{i\in \Gamma_0^+} \Delta v_i - \sum_{i\in \Gamma_1^-} \Delta v_i
    - \boldsymbol{v} \cdot \mathcal{KP}(\boldsymbol{v}, \boldsymbol{w}, \rho) + \max(\Psi^+, \Psi^-).
    \nonumber
\end{equation}
If CIS $\le 0$, the theorem indicates that the predicted importance score $\hat{\boldsymbol{v}}$ is guaranteed to induce the true KP solution $\boldsymbol{y}^*$.
If CIS $> 0$, $\hat{\boldsymbol{v}}$ does not guarantee to induce the KP solution, but a lower CIS would indicate a solution closer to $\boldsymbol{y}^*$ than a higher one.
That is, a smaller CIS indicates a stronger \textit{confidence} in satisfying the inequality, indicating a higher chance for the solution of the perturbed KP to be identical to that of the original KP.

Unlike F1 that simply measures how close the predicted summary is to the ground truth, our CIS accounts for the \textit{confidence} that the predicted score will exactly induce the true summary.
This is particularly important in video summarization, since the importance score itself is subjective and often noisy, requiring robust selection of frames. 

\vspace{0.1cm} \noindent
\textbf{Weighted Inclusion Ratios.}
Analyzing the sensitivity of the KP optimum to the perturbation of the importance score, \citet{belgacem2008sensitivity} established bounds of the perturbed profits to retain the original optimum.
These intervals are
\begin{align}
    \boldsymbol{y}^*_i &= 1 \rightarrow \Delta v_i \in I^1_i = \left[\max\left(\Delta_i^-, w_i \max_{k\in\Gamma}{\mu_k} - p_i\right), +\infty \right) \nonumber
    \\
    \boldsymbol{y}^*_i &= 0 \rightarrow \Delta v_i \in I^0_i = \left(-\infty, \min\left(\Delta_i^+, p_i - w_i \min_{k\in\Gamma}{\mu_k}\right)\right], \nonumber
\end{align}
where $\Delta_i^+ = \boldsymbol{v} \cdot (\mathcal{KP}(\boldsymbol{v}, \boldsymbol{w}, \rho) - \mathcal{KP}(\boldsymbol{v} - v_i \hat{e}_i, \boldsymbol{w}, \rho - w_i/N),  \Delta_i^- = \boldsymbol{v} \cdot (\mathcal{KP}(\boldsymbol{v} - v_i \hat{e}_i, \boldsymbol{w}, \rho) - \mathcal{KP}(\boldsymbol{v}, \boldsymbol{w}, \rho))$, and $\mu_k$ denotes the critical ratio of $k \in \Gamma$: $\mu_k = \mathcal{CR}([N] \setminus \{ k\}, \rho - \mathbf{1}_{\{y_k^*=0\}} \cdot w_k/N)$, with $\mathcal{CR}(A, \gamma) = {v_s}/{w_s}$ and $s = \min \{l| \sum_{i=1}^l v_s/w_s > \rho \}$ is the index of the critical item.
The items in $A$ are considered in descending order of ${v_s}/{w_s}$. To sum up, if the perturbed profit resides in the interval $v_i' \in I^{\boldsymbol{y}_i^*}_i$, it is guaranteed that the solution of perturbed KP is identical with that of the original KP.

We propose \textbf{Inclusion Ratio (IR)} to estimate how much the predicted scores fall into the safe bounds; for the $i$-th importance score, $\text{IR}_i \equiv |\{ \Delta v_i \in I^{y^*_i}_i \mid i \in [N] \}| / N$. 
This ratio reflects the proportion of the predicted scores that retain the KP solution.
We propose the \textbf{Weighted average of the Inclusion Ratios (WIR)} as a metric to measure the proximity of the predicted score to the ground truth: $\text{WIR} \equiv \sum_{i=1}^{n} \left(\frac{w_i}{\sum_{j=1}^{n} w_j} \text{IR}_i\right)$, where $w_i$ denotes the segment length of each shot.

\vspace{0.1cm} \noindent
\textbf{Weighted Sum of Errors (WSE).}
Additionally, we compare a weighted sum of the errors $\sum_i w_i \Delta p_i$, where $w_i$ is the segment length of each clip. 

\vspace{0.1cm} \noindent
\textbf{Comparison.}
\cref{tab:comparison_metrics} compares several competitive methods using these new metrics.
Our SummDiff achieves the best performance in all metrics, indicating that the summaries generated by our method are closer to the optimal knapsack result, under the theoretical analysis in literature.

\vspace{0.1cm} \noindent
\textbf{Discussion.}
Our newly proposed metrics, WIR and CIS, evaluate the quality of the generated importance scores in two aspects.
WIR measures how many importance scores are individually trustworthy, weighted by the duration.
Specifically, WIR first computes the safe interval $I_i$ for each importance score such that the true summary remains unchanged.
Then, given the predicted importance scores $\hat{\mathbf{v}}$, it measures how many of them fall within their corresponding safe intervals $\hat{v}_i \in I_i$.

On the other hand, CIS quantifies the risk that the predicted importance scores will lead to a different summary than the one generated using the true importance scores.
In other words, CIS evaluates the overall chance how likely the predicted score vector $\hat{\mathbf{v}}$, as a whole, is to give a summary different from the true one. 
With the analysis of knapsack optimality, these metrics would better measure the accuracy of predicted importance score, considering the fidelity of final summary \textit{after} knapsack to the ground truth, than F1 or ranking-based metrics.

\begin{table}[t]
    \vspace{-0.1cm}
    \centering
    \footnotesize
    \renewcommand{\tabcolsep}{6pt}
    \begin{tabular}{l|cc|ccc}
    \toprule
    Method & $\tau$ $\uparrow$ & $\rho$ $\uparrow$ & CIS $\downarrow$ & WIR $\uparrow$ & WSE $\downarrow$ \\ 
    
    \midrule
    Uniform-Random & 0.000 & 0.000 & 9.27 & 0.45 & 46.82 \\
    \midrule
    SL-module~\cite{xu2021cross} & 0.060 & 0.088 & 6.83 & 0.56 & 30.09 \\ 
    CSTA~\cite{son2024csta}      & 0.128 & 0.185 & 6.23 & 0.57 & 25.76 \\ 
    PGL-SUM~\cite{apostolidis2021combining}   & 0.097 & 0.141 & 6.14 & 0.58 & 26.22 \\ 
    VASNet~\cite{fajtl2019summarizing}    & 0.069 & 0.102 & 6.25 & 0.59 & 26.79 \\ 
    A2Summ~\cite{he2023align} & 0.121 & 0.172 & 6.65 & 0.55 & 30.55 \\
    
    \midrule
    \textbf{SummDiff (Ours)} & \textbf{0.175} & \textbf{0.238}  & \textbf{5.96} & \textbf{0.61} & \textbf{25.24}  \\ 

    \bottomrule
    \end{tabular}
    \vspace{-0.2cm}
    \caption{Evaluation using our new metrics proposed in \cref{sec:analysis}.}
    \label{tab:comparison_metrics}
    \vspace{-0.3cm}
\end{table}


\subsection{Ablation Study}
\label{sec:exp:ablation}

\begin{table}
  \centering
  \footnotesize
    { 
      \begin{tabular}{l|cc}
        \toprule
        Module & $\tau$ $\uparrow$ & $\rho$ $\uparrow$ \\ \midrule
        Encoder Only & 0.071 & 0.104 \\
        (+) Video Importance Denoiser & 0.079 & 0.116 \\
        (+) Learnable Embedding & 0.086 & 0.124 \\
        (+) AdaLN ($\boldsymbol{\tau}$) & 0.145 & 0.204 \\
        (+) AdaLN ($\boldsymbol{\tau}, \boldsymbol{\Phi}$) 1 layer & 0.171 & 0.232 \\
        \quad (-) Quantization & 0.125 & 0.174 \\
        (+) Classifier-free Guidance (CFG)~\cite{ho2022classifier} & 0.175 & 0.238 \\ 
        (+) Self-attention Guidance (SAG)~\cite{hong2023improving} & 0.177 & 0.239 \\ \bottomrule
      \end{tabular}}
      \vspace{-0.2cm}
    \caption{Effect of individual components}
    \label{tab:ablation}
    \vspace{-0.7cm}
  \hfill
\end{table}

We report ablation study on Mr.~HiSum~\cite{sul2024mr} in \cref{tab:ablation}, comparing performance adding a component step-by-step.
Starting from a two-layer transformer encoder predicting the importance score, adding our denoiser with a naively summed query $\mathbf{Q}_t = \mathcal{C}(\mathbf{u}_t) + \boldsymbol{\tau} + \boldsymbol{\Phi}$ improves the Kendall's $\tau$ (0.071 $\rightarrow$ 0.079).
We observe further slight improvement (0.079 $\rightarrow$ 0.086) when we replace the fixed random score embeddings with learnable ones (Learnable Embedding).
Introducing AdaLN transformation~\cite{peebles2023scalable} on the time ($\boldsymbol{\tau}$) and positional ($\boldsymbol{\Phi}$) encodings further improves $\tau$ to 0.145 and 0.171, respectively.
As we adopt the quantization idea, grouping nearby scores into a unified embedding, in \cref{sec:method:visd} with this change, we experiment without it (using a sinusoidal function~\cite{vaswani2017attention} instead), and observe significant performance drop (0.171 $\rightarrow$ 0.125). 
This confirms that quantization is superior to treating the score as a continuous scalar.
On top of this, we combine our model with two widely-known ideas, classifier-free guidance (CFG) \cite{ho2022classifier} and self-Attention guidance (SAG)~\cite{hong2023improving}, for further improvement. See \cref{appendix:sag} for details.

\begin{table}
  \centering
  \footnotesize
  \renewcommand{\tabcolsep}{4pt}
  \begin{tabular}{l|cc}
    \toprule
    Model & $\tau$ $\uparrow$ & $\rho$ $\uparrow$ \\
    \midrule
    CSTA~\cite{son2024csta} & 0.128 & 0.185 \\
    SummDiff (1 step) & 0.170 & 0.234 \\
    SummDiff (10 steps) & 0.175 & 0.238 \\
    SummDiff (100 steps) & \textbf{0.182} & \textbf{0.245} \\
    \bottomrule
  \end{tabular}
  \vspace{-0.2cm}
  \caption{Performance with various numbers of DDIM steps}
  \label{tab:perf_step1}
  \vspace{-0.3cm}
\end{table}

\vspace{0.1cm}
\noindent\textbf{Number of DDIM Steps.} \cref{tab:perf_step1} suggests that SummDiff outperforms CSTA~\cite{son2024csta}, the best-performing baseline, even with a single DDIM \cite{song2020denoising} step. 
Beyond this, SummDiff further improves its performance with more DDIM steps.

\vspace{0.1cm} \noindent
\textbf{Effect of Quantization Strength.}
Solving a KP usually has a unique solution if the profits $v_i$ are continuous.
As quantization unifies importance score values within the same interval to a single embedding, it may lead to multiple solutions to $\mathcal{KP}(\tilde{\boldsymbol{v}}, \boldsymbol{w}, \rho)$, where $\tilde{\boldsymbol{v}}$ is a quantized version of $\boldsymbol{v}$. 

To analyze this effect, we measure the average number of solutions $ \mathbb{E}_{\boldsymbol{v}_1 \sim U(0,1)^{\otimes N}} [|\{ \boldsymbol{y}^* \in \{0, 1\}^N| \boldsymbol{y}^* = \mathcal{KP}(\tilde{\boldsymbol{v}}, \boldsymbol{w}, \rho \\ )\}| ]$ while varying $K$, taking expectation over $\boldsymbol{v}$, which is a value vector sampled from a joint uniform distribution.
We also consider the average $L_1$ difference between two summaries, $\mathbb{E}_{\boldsymbol{v}_1 \sim U(0,1)^{\otimes N}} [\Delta \boldsymbol{y}^*] \equiv \mathbb{E}_{\boldsymbol{v}_1 \sim U(0,1)^{\otimes N}, \boldsymbol{v}_2 \sim \mathcal{N}(\boldsymbol{v}_1, \frac{1}{K} \mathbf{I}_N) }[\|\mathcal{KP}(\tilde{\boldsymbol{v}}_1, \boldsymbol{w}, \rho) - \mathcal{KP}(\tilde{\boldsymbol{v}}_2, \boldsymbol{w}, \\ \rho)\|_1]$, obtained by solving the KP with two similar but distinct value vectors, \textit{i.e.}, $\boldsymbol{v}_2 = \boldsymbol{v}_1 + \Delta \boldsymbol{v}$, where $\Delta \boldsymbol{v} \sim \mathcal{N}(0, \frac{1}{K} \mathbf{I}_N)$.


\begin{figure}
  \centering
  \includegraphics[width=0.39\textwidth]{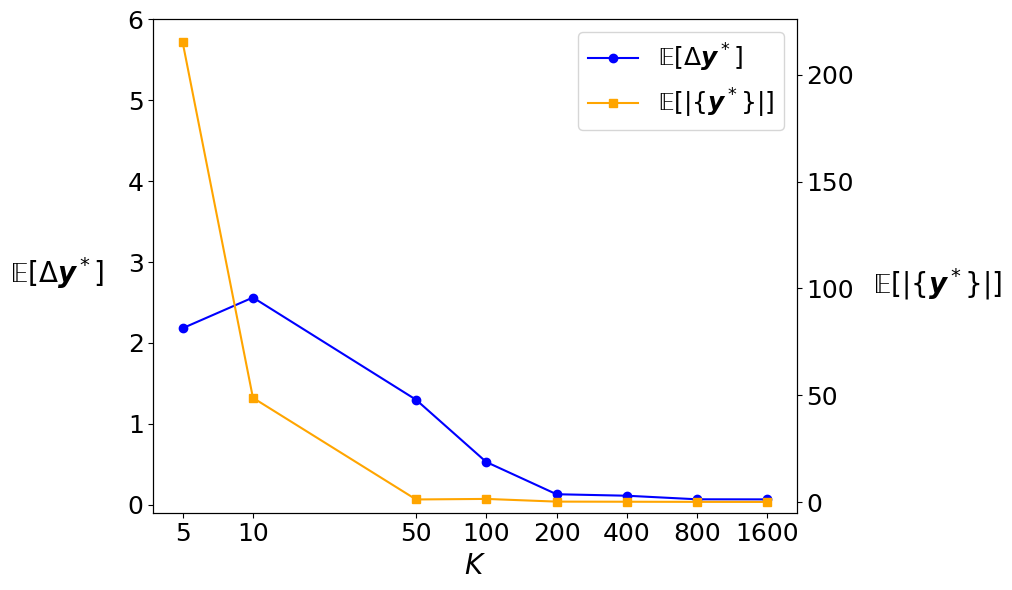}
  \vspace{-0.3cm}
  \caption{\textbf{Measured $\mathbb{E}|\{ \boldsymbol{y}^* \}|$ and $\mathbb{E} [\Delta \boldsymbol{y}^*]$ with respect to the quantization strength $K$}. A larger $K$ is associated with a smaller number of allowed solutions for the given KP.}
  \label{fig:quan}
  \vspace{-0.6cm}
\end{figure}



We numerically analyze the relation between these two values, which indicate the scale of multiplicity of the solutions to the KP, and the quantization strength $K$. As shown in \cref{fig:quan}, both the number of solutions $|\{ \boldsymbol{y}^* \}|$ and the summary deviation $\Delta \boldsymbol{y}^*$ decrease with larger $K$, suggesting more accurate summaries. 

To sum up, increasing the quantization strength $K$ reduces the number of multiple optimal solutions, leading to a more accurate summary.
However, an excessively higher $K$ leads to insufficient samples per bin, degrading the overall performance. 
See \cref{appendix:ablation_appendix} for more ablation studies.



\section{Related Work}
\label{sec:related}

\noindent\textbf{Video Summarization.}
Early models primarily rely on heuristic unsupervised learning \cite{zhang1997integrated, ngo2003automatic, mundur2006keyframe, khosla2013large, lu2013story, hong2009event, mahasseni2017unsupervised, ma2002user, de2011vsumm, song2015tvsum, chu2015video} to select important or diverse frames, struggling with  generalization.
Benefiting from annotated datasets, supervised methods \cite{potapov2014category, khosla2013large, panda2017weakly, rochan2019video, cai2018weakly} have emerged.
DSNet \cite{zhu2020dsnet} and iPTNet \cite{jiang2022joint} improve keyframe selection using frame-level annotations, and SL-Module \cite{xu2021cross} captures high-level features.
With the advent of deep learning, RNNs \cite{zhang2016video,zhao2017hierarchical,zhao2018hsa,zhao2020tth} and attention mechanism \cite{huang2021gpt2mvs, bilkhu2019attention, li2022video, terbouche2023multi, zhang2023vss, zhu2022relational, li2021exploring, apostolidis2022summarizing, jung2020global} have been adopted to capture temporal dependencies in videos.
Particularly, VASNet \cite{fajtl2019summarizing} and PGL-SUM \cite{apostolidis2021combining} capture both local and global frame dependencies using self-attention.
DMASum \cite{wang2020query} introduces mixture of attention layer mitigate the key Softmax Bottleneck.
SUM-GAN \cite{mahasseni2017unsupervised} and AC-SUM-GAN \cite{apostolidis2020ac} leverage generative adversarial networks. 

Multimodal approaches \cite{zhao2022hierarchical, hsu2023video} integrate audio or textual data to video summarization.
Multimodal transformers are adopted to link frames with corresponding captions, improving context-aware summaries, by CLIP-It \cite{narasimhan2021clip}, MSVA \cite{ghauri2021supervised}, SSPVS \cite{li2023progressive}, and A2Summ \cite{he2023align}.
CSTA \cite{son2024csta} addresses computational complexity by using CNN-based spatiotemporal attention for efficient frame selection. Unlike these deterministic summarizers, our approach generates multiple plausible summaries by capturing the distribution of video summaries with various perspectives.


\vspace{0.1cm} \noindent
\textbf{Diffusion for Video Tasks.}
Diffusion model have emerged as a groundbreaking tool in high-quality image generation \cite{ho2020denoising, song2019generative, saharia2022photorealistic, song2020improved, nichol2021glide, ryu2025, Hahm2024}.
Recently, they are extended to video generation \cite{ho2022video, ho2022imagen, esser2023structure, blattmann2023align, chen2024videocrafter2, gupta2025photorealistic, liang2025movideo}, joint video and audio generation \cite{ruan2023mm,Li2024steer,Su2024}, video editing \cite{ceylan2023pix2video, chai2023stablevideo, tu2024motioneditor}, video inpainting \cite{zhang2024avid} and prediction \cite{hoppe2022diffusion, zhang2024extdm}.
Furthermore, they are applied to video understanding tasks like moment retrieval \cite{li2024momentdiff, luo2024generative}, video object segmentation \cite{zhu2024exploring, zhou2023quality}, action segmentation \cite{liu2023diffusion}.

\section{Summary}
\label{sec:Conclusion}

Video summarization is inherently subjective since people have different perspectives of a good summary.
We suggest a generative viewpoint of this task where the model learns the distribution of good summaries, in contrast to the traditional approach of aggregated importance score regression.
Our proposed SummDiff, adopting diffusion models for video summarization for the first time, is able to generate multiple good summaries conditioned on the input video.
Our model not only outperforms other baselines but also demonstrates the ability to generate accurate summaries customized to the individual annotators.
We further propose additional metrics that measure the quality of the predicted importance scores through an insight from the actual summary generation using knapsack.



\section*{Acknowledgments}
This work was supported by Samsung Electronics (IO240512-09881-01), Youlchon Foundation, NRF grants (RS-2021-NR05515, RS-2024-00336576, RS-2023-0022663) and IITP grants (RS-2022-II220264, RS-2024-00353131) by the government of Korea.

{
    \small
    \bibliographystyle{ieeenat_fullname}
    \bibliography{main}
}

\clearpage
\setcounter{page}{1}
\maketitlesupplementary
\appendix

\pagenumbering{roman}
\renewcommand\thesection{\Alph{section}}
\renewcommand\thetable{\Roman{table}}
\renewcommand\thefigure{\Roman{figure}}
\setcounter{section}{0}
\setcounter{table}{0}
\setcounter{figure}{0}

\section{Additional Ablation Studies}
\label{appendix:ablation_appendix}
\vspace{0.1cm}

\noindent\textbf{Effect of Quantization.}
We explore various number of segments ($K$) for the codebook $\mathcal{C}$, uniformly splitting the score range $[0, 1]$.
According to \cref{tab:k}, the best performance is achieved around 200 to 400.
With $K < 200$, the performance degrades because it limits the ability of our model to distinguish different scores, treating a wide range of scores with the same embedding.
With too large $K$, on the other hand, the model would suffer from the lack of samples per each bin, degrading the performance.

\begin{table}[h]
\centering
    {\footnotesize
    \begin{tabular}{@{}ccc@{}}
        \toprule
        $K$ & $\tau$ & $\rho$ \\ \midrule
        5    & 0.145 & 0.200 \\
        10   & 0.147 & 0.202 \\
        50   & 0.147 & 0.202 \\
        100  & 0.171 & 0.235 \\
        200  & 0.175 & 0.238 \\
        400  & 0.173 & 0.237 \\
        800  & 0.172 & 0.235 \\
        \bottomrule
    \end{tabular}}
    \vspace{-0.2cm}
    \caption{Effect of quantization strength $K$}
    \label{tab:k}
\end{table}

\begin{figure}[h]
    \centering
    \vspace{-5pt}
    \includegraphics[width=0.47\textwidth]{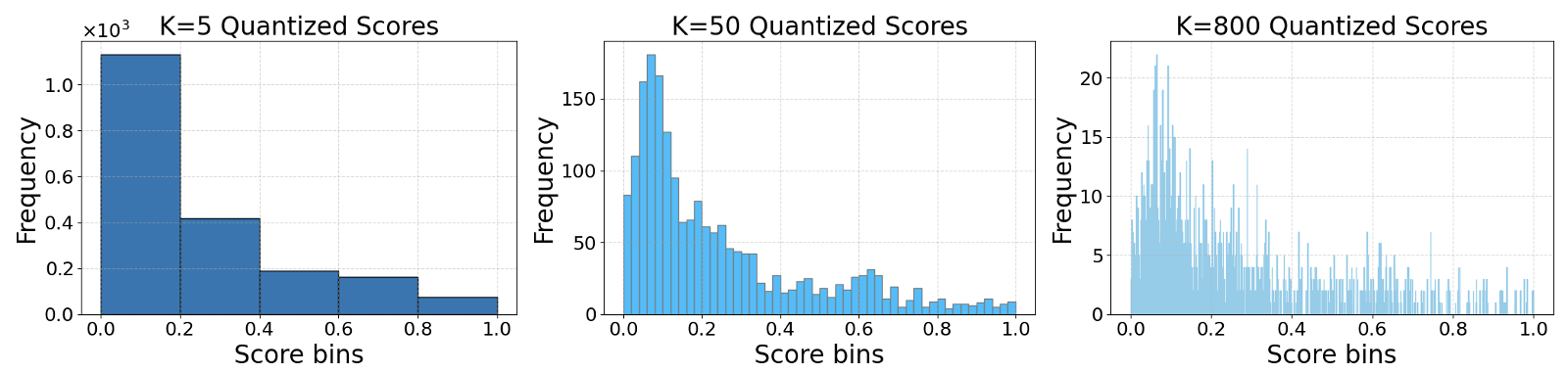}
    \caption{Histogram of quantized importance scores for varying $K$ on a subset of the Mr.~HiSum dataset.}
    \label{fig:q_bins}
    \vspace{-0.4cm} 
\end{figure}

\noindent\textbf{Visualization of Histograms from Quantization.}
To further visualize the effect of quantization with varying $K$, we illustrate the distribution of quantized scores for different values of $K$ using histograms, and count the number of scores falling into each bin on a subset of Mr.~HiSum dataset shown in \cref{fig:q_bins}. When $K$ is too small (left), quantization becomes too coarse, collapsing diverse scores into the same bin and causing more multiple optimal solutions observed in \cref{fig:quan}. When $K$ is too large (right), bins become too sparse, leading to unstable estimates with insufficient samples per bin. Both extremes hurt performance. As shown in \cref{appendix:ablation_appendix}, the best results are achieved when $K$ strikes a balance between granularity and robustness.

\vspace{0.1cm}
\noindent\textbf{Inference Time.}
The iterative sampling process in generative diffusion might raise concerns about slower inference time.
We report the average inference time of top-performing models under the same condition in \cref{tab:inference_time}.
SummDiff (1 step) takes comparable time to others including CSTA \cite{son2024csta}, and \cref{tab:perf_step1} confirms that this setting outperforms other baselines.
In short, SummDiff (1 step) achieves moderately fast inference time with a reasonably strong summarization performance, effectively balancing them.

\begin{table}[h]
\centering
    {\footnotesize
    \begin{tabular}{l|c}
        \toprule
        Model & Inference Time (ms) \\
        \midrule
        CSTA~\cite{son2024csta}          & 11.70 $\pm$ 0.11 \\
        PGL SUM~\cite{apostolidis2021combining}       & 19.32 $\pm$ 0.50 \\
        SL\_Module~\cite{xu2021cross}    & 5.20 $\pm$ 0.62 \\  
        VASNET~\cite{fajtl2019summarizing}        & 1.11 $\pm$ 0.23 \\
        A2Summ~\cite{he2023align}        & 9.10 $\pm$ 2.25 \\
        \midrule
        \textbf{SummDiff} (1 step)   & 11.02 $\pm$ 1.92 \\
        \textbf{SummDiff} (10 steps)  & 49.73 $\pm$ 1.55 \\
        \bottomrule
    \end{tabular}}
    \vspace{-0.2cm}
    \caption{Comparison on inference time}
    \label{tab:inference_time}
\end{table}

\noindent\textbf{Training on aggregated scores.}
To further investigate the importance of training on individual annotator scores, we conduct additional experiments on SumMe with various number of annotations, $|\mathcal{R}| \in \{5, 10, \text{All} \}$.
We train a model on a randomly selected set of annotations of size $|\mathcal{R}|$, either on individual annotations or on their aggregated scores.
As seen in the table, training on individual scores consistently outperforms.
This supports our claim that using individual scores aligns well with our generative approach and leads to higher-quality summaries.

\begin{table}[htbp]
\centering
{\footnotesize
\begin{tabular}{c|c|cc}
\hline
$|\mathcal{R}|$ & Training & $\tau$ & $\rho$ \\
\hline
5   & Agg & 0.130     & 0.144     \\
    & Ind & \textbf{0.211}   & \textbf{0.236}     \\
\hline
10  & Agg & 0.145     & 0.161     \\
    & Ind & \textbf{0.227}     & \textbf{0.253}    \\
\hline
All & Agg & 0.176  & 0.196  \\
    & Ind & \textbf{0.256} & \textbf{0.285} \\
\hline
\end{tabular}}
\caption{Performance comparison on SumMe when training with aggregated (Agg) versus individual (Ind) annotator scores, across different numbers of annotations $|\mathcal{R}| \in \{5, 10, \text{All} \}$.}
\label{appendix:tab:agg}
\end{table}



\section{Classifier-free Guidance and Self-attention Guidance}
\label{appendix:sag}
We integrate two widely-known ideas for further improvement of SummDiff. 
First, we adopt the classifier-free guidance (CFG) \cite{ho2022classifier}, $\Tilde{f}_{\theta}(\mathcal{C}(\mathbf{u}_t), t, \mathbf{Z}) = (1 + w) f_{\theta}(\mathcal{C}(\mathbf{u}_t), t, \mathbf{Z}) - w f_{\theta}(\mathcal{C}(\mathbf{u}_t), t, \varnothing)$,
where $\varnothing$ is the null (black) video, and $w$ determines the extent to which unconditioned information is used at inference.

Second, we add self-Attention guidance (SAG)~\cite{hong2023improving}, leveraging the intermediate self-attention maps from diffusion models to improve stability.
It works by selectively blurring the areas that the diffusion models focus on during each step, using an adversarial approach to adjust and guide the model’s attention as it progresses. Specifically,
\begin{align}
    \Tilde{f}_{\theta}(\mathcal{C}(\mathbf{u}_t), t, \mathbf{Z}) &= \Tilde{f}_{\theta}(\mathcal{C}(\mathbf{u}_t), t, \mathbf{Z}) \\
    &+ (1 + s)(\Tilde{f}_{\theta}(\mathcal{C}(\mathbf{u}_t), t, \mathbf{Z})) -  \Tilde{f}_{\theta}(\mathcal{C}(\hat{\mathbf{u}}_t), t, \mathbf{Z})
    \nonumber
\end{align}
where $A_t$ denotes the attention map, $M_t = \mathbbm{1}(A_t > \psi)$ is a binary mask indicating where $A_t$ exceeds a threshold $\psi$, and $\odot$ is the Hadamard product. 
The intermediate reconstruction $\mathcal{C}(\hat{\mathbf{u}}_t)$ selectively combines the original signal $\mathcal{C}(\mathbf{u}_t)$ and its noised version $\mathcal{C}(\mathbf{\Tilde{u}}_t)$ based on $M_t$ by
\begin{equation}
    \mathcal{C}(\hat{\mathbf{u}}_t) = (1-M_t)\odot \mathcal{C}(\mathbf{u}_t) + M_t\odot \mathcal{C}(\Tilde{\mathbf{u}}_t),
\end{equation}
where $\mathcal{C}(\Tilde{\mathbf{u}}_0)$ is obtained by convolving $\mathcal{C}(\mathbf{u}_0)$ with a Gaussian kernel $G_{\sigma}$. Finally, $\mathcal{C}(\mathbf{\hat{u}}_0)$ is computed by
\begin{equation}
    \mathcal{C}(\mathbf{\hat{u}}_0)=(\mathcal{C}(\mathbf{u}_t)-\sqrt{1-\Bar{\alpha}}_tf_{\theta}(\mathcal{C}(\mathbf{u}_t), t, \mathbf{Z})/\sqrt{\Bar{\alpha}_t}.
\end{equation}
$\mathcal{C}(\Tilde{\mathbf{u}}_t)$ is obtained by diffusing with noise $f_{\theta}(\mathcal{C}(\mathbf{u}_t), t, \mathbf{Z})$ from ${\mathcal{C}(\mathbf{\Tilde{u}}_0)}$.

In \cref{tab:ablation}, we observe extra gains with these, achieving 0.177 of $\tau$.
A similar pattern of gradual improvement is observed with Spearman's $\rho$, as seen in \cref{tab:ablation}.

\section{Qualitative Results on Contours}
\label{sec:appendix:contour}

\begin{figure}
  \begin{center}
  \includegraphics[width=\linewidth]{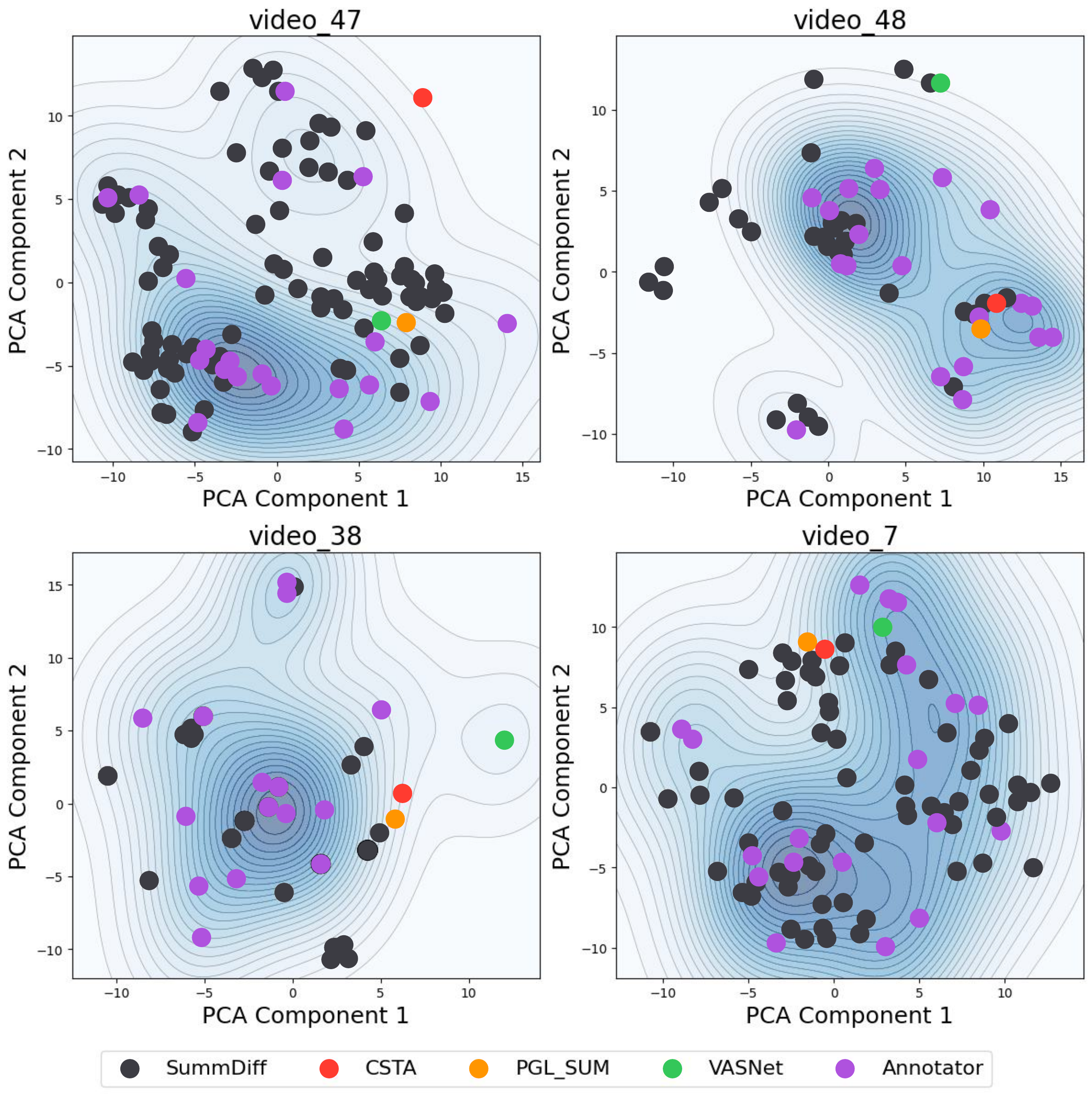}
  \end{center}
  \caption{Distribution of ground truth and generated summaries of selected videos}
  \label{fig:qualitative_appendix}
\end{figure}

We provide additional examples of generated summaries in \cref{fig:qualitative_appendix}.
The summaries annotated by human raters (\mycirc[cyan] Annotator) and those generated by ours (\mycirc[darkgray] SummDiff) and three baselines (\mycirc[red] CSTA, \mycirc[green] PGL-SUM, and \mycirc[blue] VASNet) are projected to 2D using PCA.
The annotated summaries are visualized using a blue contour map created through Gaussian kernel density estimation. This highlights that SummDiff produces a diverse range of summaries that closely match those provided by human annotators. For instance, in the top-right plot for video 48, human evaluators identified three distinct ways to summarize the content. SummDiff successfully generates summaries that capture all three variations. In contrast, the baseline models typically produce a single, less accurate summary, failing to capture the variability seen in human-generated summaries. Similar trends are observed across the other three plots.

\section{Evaluation Results with Standard Deviation}
\label{appendix:full_table}

\cref{tab:leaderboard_std} shows the full evaluation results from \cref{tab:new_leaderboard} with standard deviation.
As seen in the table, our method is superior to all other baselines statistically significantly.

\section{Implementation Details}
\label{sec:impl_detail_appendix}
We uniformly sample frames from each video at 1 fps matching the ground truth label provided in the Mr.~HiSum dataset.
For TVSum and SumMe, the videos are subsampled to 2 fps as in~\cite{son2024csta,ghauri2021supervised,hsu2023video,li2023progressive,mahasseni2017unsupervised,zhang2016video}.
For Mr.~HiSum, we employ 2 transformer layers for visual encoding and 2 additional layers for denoising.
Each layer has a hidden size of 256, 8 attention heads, and feed-forward network with a dimensionality of 1024. For TVSum and SumMe we adopt 1D convolution layer followed by an MLP for the decoder due to the small dataset size.
We use AdamW optimizer~\cite{loshchilov2017decoupled} with cosine annealing~\cite{touvron2021training}, gradually reducing the learning rate from $5\cdot 10^{-5}$ initially. We also used EMA decay of 0.999 for Mr.~HiSum.
We set the batch size to 256, and train up to 200 epochs. $\epsilon$ is set to $10^{-3}$ when clipping $\boldsymbol{s}_0$.
For inference, we take 10 DDIM~\cite{song2020denoising} steps by default. 
We conduct all experiments on a single NVIDIA A5000 GPU.

We investigate various batch sizes from the set $\{32, 64, 256\}$ and determined that a batch size of 256 yielded the best performance for SummDiff model. We apply tuning procedures consistently across all models. For SumMe and TVSum, batch size of 20 and 40 yielded the best result respectively.

Furthermore, we experiment with learning rates and weight decay within the range of \{$10^{-3}$, $5\cdot 10^{-4}$, $10^{-4}$, $5\cdot10^{-5}$, $10^{-5}$\}.
Learning rate of $5\cdot 10^{-5}$ and weight decay of $5\cdot10^{-4}$ is found to be optimal for our SummDiff model, and similar tuning procedure has been applied to other models.

\clearpage

\begin{table*}
    \centering
    \footnotesize
    \renewcommand{\tabcolsep}{5pt}
    {\begin{tabular}{lc|cc|cc}
    \toprule
    Model & Year & $\tau$ $\uparrow$ &  $\rho$ $\uparrow$ & MAP{\tiny$\rho$ = 50\%} $\uparrow$ & MAP{\tiny $\rho$ = 15\%} $\uparrow$ \\
    \midrule
    SUM-GAN \cite{mahasseni2017unsupervised} & 2017 & 
    0.067\scriptsize{$\pm$0.018} &
    0.095\scriptsize{$\pm$0.023} & 
    59.50\scriptsize{$\pm$0.16} & 24.30\scriptsize{$\pm$0.19} \\
    
    VASNet~\cite{fajtl2019summarizing}  & 2019 &
    0.069\scriptsize{$\pm$0.000} & 0.102\scriptsize{$\pm$0.000} & 
    58.69\scriptsize{$\pm$0.30} & 25.28\scriptsize{$\pm$0.40} \\
    
    AC-SUM-GAN \cite{apostolidis2020ac} & 2020 & 
    0.012\scriptsize{$\pm$0.003} &
    0.018\scriptsize{$\pm$0.003} & 
    56.40\scriptsize{$\pm$0.06} & 21.70\scriptsize{$\pm$0.08} \\
    
    SL-module \cite{xu2021cross} & 2021 & 
    0.060\scriptsize{$\pm$0.002} & 0.088\scriptsize{$\pm$0.003} & 
    58.63\scriptsize{$\pm$0.13} & 24.95\scriptsize{$\pm$0.13} \\
    
    PGL-SUM~\cite{apostolidis2021combining} & 2021 & 
    0.097\scriptsize{$\pm$0.001} & 0.141\scriptsize{$\pm$0.001} & 
    61.60\scriptsize{$\pm$0.14} & 27.45\scriptsize{$\pm$0.15} \\
    
    iPTNet \cite{jiang2022joint} & 2022  &
    0.020\scriptsize{$\pm$0.003} & 0.029\scriptsize{$\pm$0.004} & 
    55.53\scriptsize{$\pm$0.25} & 22.74\scriptsize{$\pm$0.13} \\
    
    A2Summ \cite{he2023align} & 2023 &
    0.121\scriptsize{$\pm$0.001} &
    0.172\scriptsize{$\pm$0.001} & 
    59.18\scriptsize{$\pm$0.13} & 30.70\scriptsize{$\pm$0.21} \\
    
    CSTA ~\cite{son2024csta} & 2024 &
    0.128\scriptsize{$\pm$0.004} & 0.185\scriptsize{$\pm$0.006} & 
    62.25\scriptsize{$\pm$0.15} & 28.42\scriptsize{$\pm$0.33} \\
    
    \midrule
    
    \textbf{SummDiff}  & Ours &
    \textbf{0.175}\scriptsize{$\pm$0.005} &
    \textbf{0.238}\scriptsize{$\pm$0.004} & 
    \textbf{65.44}\scriptsize{$\pm$0.19} & 
    \textbf{33.83}\scriptsize{$\pm$0.44} \\
    
    \bottomrule
    \end{tabular}}
  \vspace{-0.2cm}
  \caption{\textbf{Comparison of models trained with Mr.~HiSum}}
  \label{tab:leaderboard_std}
\end{table*}

\section{Additional Qualitative Results}
\label{appendix:qual}
As illustrated in \cref{fig:demo}, we visualize the summary of videos generated by various models. Specifically, CSTA~\cite{son2024csta}, A2Summ~\cite{he2023align}, PGL-SUM~\cite{apostolidis2021combining}, VASNet~\cite{fajtl2019summarizing} is compared against our model, SummDiff. All videos are selected from the test set of TVSum from 5-fold cross validation experiment. Two different summary annotations are visualized in the first and second rows using black and red boxes. The results demonstrate that SummDiff predicts both more accurate and various summaries compared to other baseline models.

\begin{figure*}[t]
  \centering
  \includegraphics[width=0.94\textwidth]{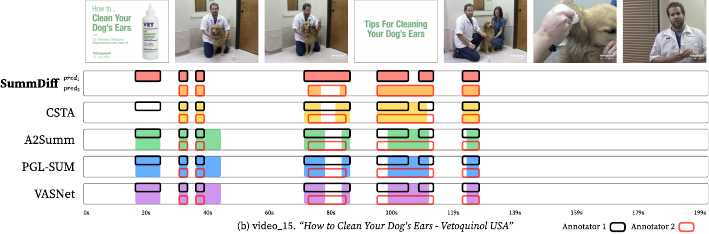}
  \caption{\textbf{Additional demonstration of video summaries generated by competing methods on a TVSum video}. Shaded parts indicate the segments selected by each method, and the two rows of edged boxes within each method indicate two different true annotations. The results clearly demonstrate the effectiveness of SummDiff in capturing multiple plausible summaries for a video.}
  \label{fig:demo_appendix}
\end{figure*}

\end{document}